\documentclass[a4paper, 10pt, conference]{ieeeconf}      
\usepackage{graphicx}
\usepackage{multirow}

\IEEEoverridecommandlockouts                              
\title{\LARGE \bf
Smile detection in the wild based on transfer learning
}

\begin{document}

\author{\parbox{16cm}{\centering
   {\large Xin Guo$^1$ and Luisa F. Polan\'{i}a$^2$ and Kenneth E. Barner$^1$}\\
    {\normalsize
    $^1$ Department of Electrical and Computer Engineering, University of Delaware, Newark, DE 19716, USA\\
    $^2$ American Family Mutual Insurance Company, Madison, WI 53783, USA}}
    \thanks{This material is based upon work supported by the National Science Foundation under Grant No. 1319598.}
}

\pagestyle{plain}
\maketitle

\begin{abstract}
Smile detection from unconstrained facial images is a specialized and challenging problem. As one of the most informative expressions, smiles convey basic underlying emotions, such as happiness and satisfaction, which lead to multiple applications, ~\textit{e.g.}, human behavior analysis and interactive controlling. Compared to the size of databases for face recognition, far less labeled data is available for training smile detection systems. To leverage the large amount of labeled data from face recognition datasets and to alleviate overfitting on smile detection, an efficient transfer learning-based smile detection approach is proposed in this paper. Unlike previous works which use either hand-engineered features or train deep convolutional networks from scratch, a well-trained deep face recognition model is explored and fine-tuned for smile detection in the wild. Three different models are built as a result of fine-tuning the face recognition model with different inputs, including aligned, unaligned and grayscale images generated from the GENKI-4K dataset. Experiments show that the proposed approach achieves improved state-of-the-art performance. Robustness of the model to noise and blur artifacts is also evaluated in this paper. 

\end{abstract}

\section{INTRODUCTION}
\label{sec1}
A smile is considered the most common human facial expression to convey emotions of joy, happiness, and satisfaction~(\cite{Ekma88}). Smile detection has multiple applications in different domains, such as human behavior analysis~(\cite{Pant07}), photo selection~(\cite{Pota09}), product rating~(\cite{Tava04}), and patient monitoring~(\cite{Yada12}). Recent longitudinal studies have used smile information from images to predict future social and health outcomes~(\cite{Abel10, Hark01, Sede12}). For example,~\cite{Sede12} showed that the smile intensity in Facebook profile pictures is correlated with satisfying social relationships and is a predictor of self-reported life satisfaction after 3.5 years. Another application is related to the smile shutter function of modern consumer cameras. In 2007, Sony released its first camera Cyber-shot DSC T200 equipped with a smile shutter function that perceives three human faces in the scene and takes a photograph if a smile is perceived. Similarly, in 2011, Samsung released its first smart phone with a smile shutter functionality, the Samsung Galaxy mini S5570. It is reported that the smile shutter in both the Sony and the Samsung devices is only capable of detecting big smiles but unable to detect slight smiles~(\cite{Yada12}). All these applications motivate the development of robust and automatic smile detection algorithms.

\begin{figure*}
  \centering
    \includegraphics[width=0.8\textwidth]{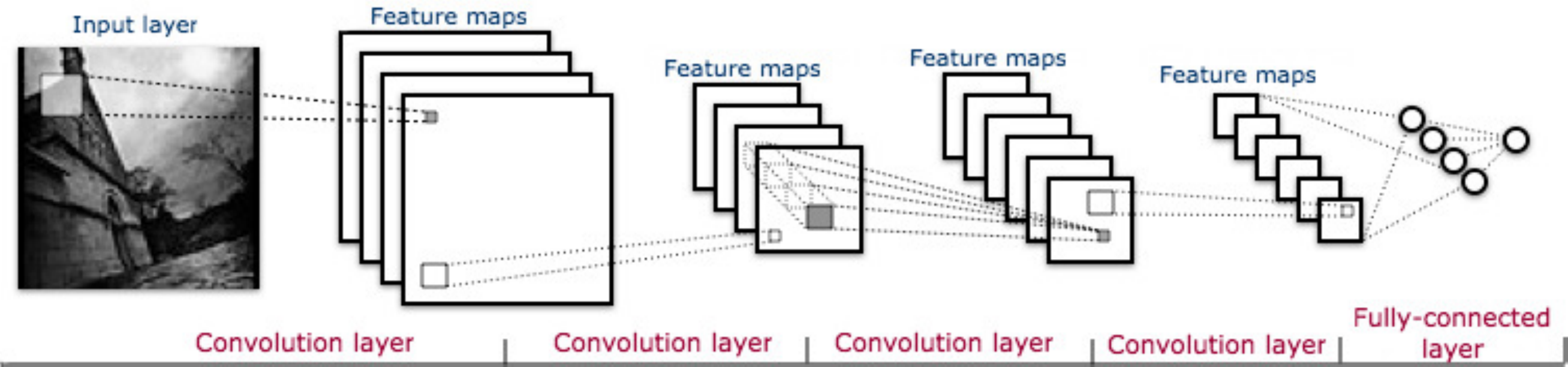}
  \caption{Example of a CNN architecture.}
  \label{CNN}
\end{figure*}

During the last two decades, the image processing and computer vision communities have developed many smile detection algorithms~(\cite{White09}). For example, in~\cite{Frei09}, local binary patterns (LBP) were used as main image descriptors for smile detection. The authors reported a classification accuracy of 90\% using support vector machines (SVM) and a small dataset of 5781 images. A smile detector based on the Viola-Jones cascade classifier was proposed in~\cite{Deni08}. The detector achieved a classification accuracy of 96.1\% on a small testing dataset of 4928 images. Although the classification accuracy was high, the employed images were mainly frontal and were captured under tightly controlled conditions. 

An important contribution to the field of automatic smile detection was introduced by~\cite{White09}. They collected the GENKI-4K database, which contains 4000 real-life face images, downloaded from publicly available Internet repositories, which have been labeled as either smiling or non-smiling by human coders. The relevance of this dataset is that it contains a large number of images that span a wide range of imaging conditions and camera models, as well as variability in ethnicity, gender, age, and background. Prior to the GENKI-4K dataset, the employed datasets for smile detection were overly constrained and led to non-generalizable results. The GENKI-4K database has become the standard dataset for evaluating smile recognition algorithms in the wild. For example,~\cite{Shan12} proposed a smile detection approach that uses intensity differences between pixels in the grayscale representation of the GENKI-4K face images as features. They used AdaBoost to choose and combine weak classifiers and reported a classification accuracy of 88\%. 


Among the latest algorithms for smile detection are the convolutional neural networks (CNN)-based algorithms, which learn hierarchical feature representations with higher level features formed by the composition of lower level features~(\cite{Lecu95}). The high classification accuracy achieved by CNNs has led them to become the state-of-the-art in smile detection in the wild. For example, in~\cite{Zhan15}, a CNN architecture of 4 convolutional layers and 1 fully-connected layer was trained from scratch for smile detection. The authors achieved a classification accuracy of 94.6\%, which is greater than any accuracy attained by previous methods on the GENKI-4K dataset. Similarly, \cite{Glau15} proposed to use a CNN for smile detection. They used model selection for choosing  the CNN parameters and used both the face and mouth regions as inputs. They used the DISFA database~(\cite{Mava13}), whose images were captured under laboratory-controlled conditions. A CNN architecture, referred to as Smile-CNN, was recently proposed by \cite{Chen16} to perform smile detection. The architecture consists of 3 convolutional layers and 1 fully-connected layer, and was trained from scratch on images from the GENKI-4K dataset. The authors attained an average classification accuracy of 92.4\% and 91.8\% using an SVM and an AdaBoost classifier, respectively. 

In this paper, we introduce a CNN-based approach that uses transfer learning to achieve a classification accuracy of 95.38\% on the GENKI-4K dataset, which is greater than that obtained by previous CNN-based smile recognition methods. Specifically, our contributions include the fine-tuning of the VGG-face model (\cite{Park15}) with images from the GENKI-4K dataset, incorporation of face alignment to enhance the performance of the CNN model, and evaluation of the model robustness to image artifacts, such as noise and blur. The motivation for fine-tuning a pre-trained model is that available datasets for smile recognition are small to train a deep neural network from scratch. Even the GENKI-4K dataset has a limited size of 4000 images, which differs from the large-scale datasets typically used to train CNNs from scratch, which are in the order of millions of images (\cite{Park15, Kriz12}).







%
%

\section{Background}
Remarkable progress has been made in image recognition in recent years, mainly due to the availability of large-scale labeled datasets, modern graphics processors, the revival of deep CNNs, and the capability of CNNs to enable transfer learning. This section briefly describes CNNs and transfer learning. 

\begin{figure*}
  \centering
    \includegraphics[width=\textwidth]{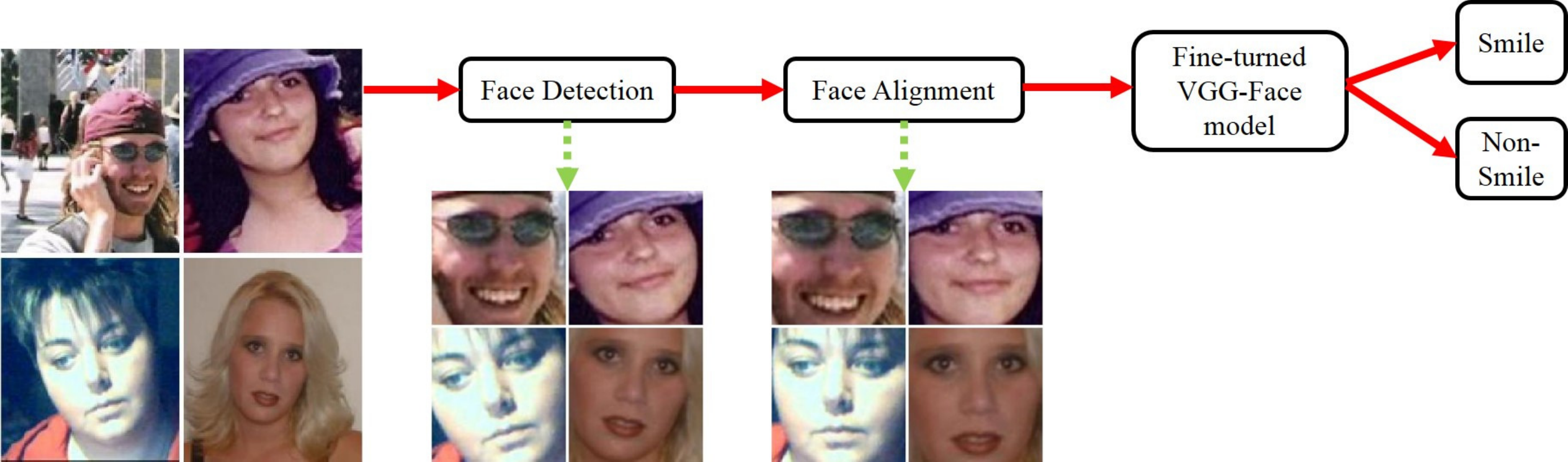}
   \caption{System diagram. Original facial images are first detected and aligned, then fed to the CNN  model for classification into two categories: smile and non-smile.}
  \label{Diagram}
\end{figure*}

\subsection{Convolutional Neural Networks}
In the context of visual recognition, a CNN is a type of feed-forward neural network that learns image features from pixels through convolutions, matrix multiplications, and nonlinear transformations, constructing a non-linear mapping between the input and the output. The lower convolutional layers extract and combine local features from the input image, and the top convolutional layers are able to learn more complicated structures by combining features from previous layers. Fully-connected layers convert the features of the top convolutional layers into a 1-dimensional vector that is categorized by a trainable classifier. CNNs are trained using backpropagation ~(\cite{Lecu15}). Fig.~\ref{CNN} illustrates, as an example, a CNN architecture of 4 convolutional layers and 1 fully-connected layer.

Feature learning for images is not a trivial problem because there are scale, orientation, and position variations for individual images. To mitigate for the high variability in the data and ensure some degree of scale, translation, and orientation invariance, CNNs combine local receptive fields, shared weights, and downsampling. Local receptive fields mean that a layer receives inputs from a set of units located in a small neighborhood in the previous layer, which allow neurons to extract primitive visual features, such as oriented edges and corners. Since units in a layer are organized in planes, weight sharing means that units within a plane share the same set of weights and perform the same operation on different parts of the image. Such planes and set of weights are usually referred to as feature maps and filter banks, respectively. 

The filtering operation performed by a feature map is equivalent to discrete convolution, which earned the CNN its name. Apart from enforcing shift-invariance, weight sharing is essential for reducing the number of trainable parameters, which otherwise may grow very rapidly for high-dimensional inputs and lead to intractable networks. Downsampling refers to the introduction of layers to reduce the resolution of the feature maps, which in turn reduces the sensitivity to small translations and distortions. A typical downsampling technique is known as max-pooling and consists of computing the maximum of a local patch of units in one feature map. A standard CNN architecture contains four types of layers: convolutional, fully-connected, activation, and pooling. 

\subsection{Transfer learning}
Two difficulties of training CNNs are the high number of needed training samples and annotations and the long time required to fully train the networks (\cite{Simo14}). There are many applications that suffer from deficit of training samples, and therefore, fully training a CNN becomes impractical for those cases. For example, medical imaging applications. 

The concept of transfer learning, as applied to visual recognition tasks, refers to transferring image representations learned with CNNs on large datasets to other visual recognition tasks with limited training data (\cite{Oqua14}). The intuition behind this idea is that convolutional layers provide generic mid-level image representations that can be transferred to new tasks. The natural hierarchical feature representation of CNNs, going from low-level features to more complex features, allows features to be shared between unrelated tasks.

Features learned from large-scale datasets,~\textit{e.g.}, ImageNet, can replace hand-crafted features in other tasks. Specifically, the process starts by removing the last fully-connected layer and then, use the rest of the CNN as a fixed feature extractor for the new dataset. Alternatively, the features learned from large-scale datasets can serve as a better weight initialization for networks being trained with limited data. This process is known as fine-tuning and can be performed in two different ways. That is, either all the layers of the CNN can be fine-tuned or the high-level portion of the CNN can be fine-tuned while some of the earlier layers can be kept fixed to prevent overfitting. The motivation for fine-tuning is that the low-level features of a CNN contain more generic features,~\textit{e.g.}, edge detectors, while higher-level features  become more specific to the given task.

\section{Methods}
This section describes the face detection and alignment methods used in our experiments to preprocess the raw facial images, and the details for fine-tuning the VGG-face model. The pipeline of the method is shown in Fig.~\ref{Diagram}.

\subsection{Preprocessing}
\label{preprocessing}
Faces are first detected using the method described in~\cite{kazemi2014one}. The motivation for using this method is that it provides facial landmarks which can be used for face alignment via a 2D affine transformation where the left and right eye corners of all the images are aligned to the same positions. The next step is to crop and rescale the face regions to $256 \times 256$ pixels. Samples of cropped and aligned face patches are shown in Fig.~\ref{Diagram}. 

Smile detection requires the modeling of subtle and localized variations between images. By eliminating some of the other type of variability in the data that is not relevant for smile detection,~\textit{e.g.}, head pose variations, face alignment is expected to help the network better learn the optimal features for smile detection.

\subsection{VGG Net}
The 16-layer VGG architecture was presented in~\cite{Simo14} as a new architecture that bids the performance of AlexNet~(\cite{Kriz12}) in the ImageNet Large-Scale Visual Recognition Challenge (ILSVRC). To this end, the authors increased the depth of AlexNet by adding more convolutional layers and used smaller convolution filters to keep the number of parameters tractable. 

The 16-layer VGG architecture is as follows. The input to VGG is a fixed-size RBG image of $224\times224$ pixels. All the convolutional layers have a fixed small receptive field which is $3\times3$. The convolution stride and the spatial padding are both fixed to $1$ pixel. Each convolutional layer is followed by a Rectified Linear Unit (ReLU) layer as in~\cite{Kriz12}. Spatial downsampling is performed through max-pooling over a $2\times2$ pixel window, with stride $2$, after $2$ or $3$ convolutional layers. A stack of 13 convolutional layers are followed by three fully-connected layers, where the first two have $4096$ channels each, and the third has a number of channels that depends on the classification task. The last layer is the soft-max layer. The activation function for each fully-connected layer is a rectified linear unit as well.

In~\cite{Park15}, the VGG-face model was presented as the result of training the 16-layer VGG architecture on a very large-scale dataset for face recognition. The dataset contained 2.6M images of 2.6K celebrities and public figures. The VGG-face model became the state-of-the-art for face recognition on the YouTube Faces dataset. 

\begin{figure}
  \centering
    \includegraphics[width=\columnwidth]{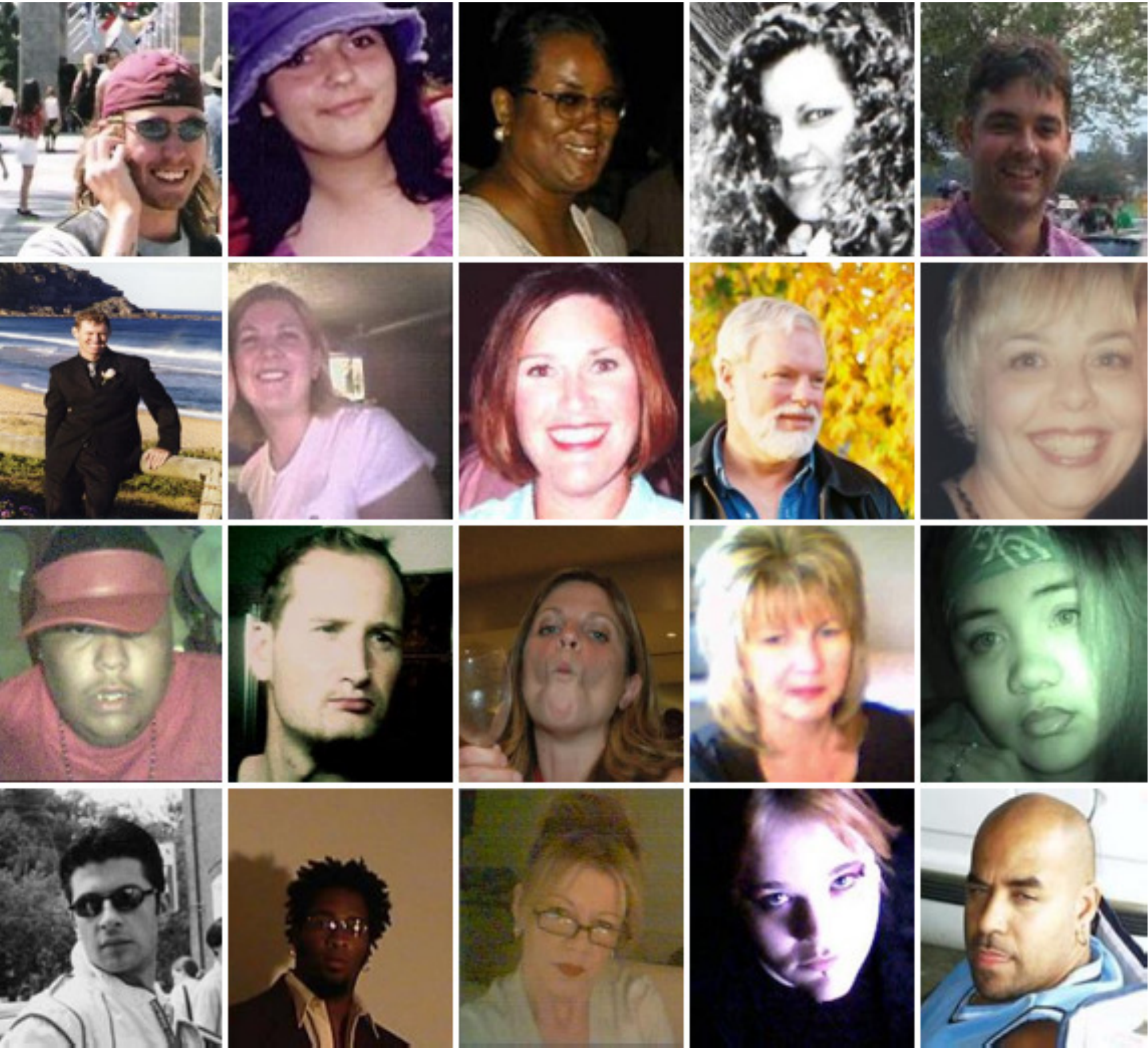}
  \caption{Examples of smiling (top two rows) and non-smiling (bottom two rows) faces in the wild. Images are from the GENKI-4K database.}
  \label{Sample}
\end{figure}

\subsection{Fine-tuning}
\label{fine-tuning}
The architecture of the VGG-face model is modified by changing the number of neurons in the last fully-connected layer to 2, indicating a binary classification having as targets smile and non-smile facial expressions. With the exception of the last fully-connected layer, the modified architecture is initialized with the VGG-face model~(\cite{Park15}), which is expected to be better than random Gaussian weights initialization since it was trained on 2.6M facial images. The last fully-connected layer is initialized with weights sampled from a Gaussian distribution of zero mean and variance $1\times 10^{-4}$. 

Because the features learned from CNN layers typically correspond to generic features, such as contours and edges, the weights of all the convolutional layers are kept the same as in the VGG-face model, while the weights of the first two fully-connected layers are fine-tuned and the last fully-connected layer is trained from scratch.

The goal is to train the model that minimizes  the average error of the final soft-max layer. The learning parameters of the model, such as learning rate and weight decay of the network, are set the same as in the VGG-face model, then gradually adjusted based on grid search. As a result, all the model learning parameters are the same as in the VGG-face model except the initial learning rate, which is scaled by a factor of 10. More precisely, the learning rate is initially set to $1 \times 10^{-3}$ and then decreased by a factor of 10 when the accuracy on the validation set stops increasing. The weight decay coefficient is set to $5 \times 10^{-4}$. Stochastic gradient descent is used to optimize the network with mini-batches of 64 samples and momentum coefficient of $0.9$. 

The training images are rescaled to $256\times256$ and randomly cropped to $224\times224$ patches to generate the input to the network. The training data is further augmented by flipping the images horizontally with 50\% probability.

\section{Experimental Results}
This section describes the experiments that validate the performance of the fine-tuned VGG-face model for smile detection using three different inputs, including aligned, unaligned and grayscale images generated from the GENKI-4K dataset. Evaluation of the models is performed in terms of classification accuracy and robustness to noise and blur artifacts. All the experiments were carried on 2 NVIDIA K40C GPUs, each with 12GB GDDR5.

\subsection{Database}
The VGG-face model is fine-tuned and tested on the GENKI-4K~(\cite{White09}) database. The images were taken by ordinary people for their own purpose, thus resulting in a wide range of imaging conditions, both outdoors and indoors, as well as variability in illumination, pose (yaw, pitch and roll parameters of the head of most of the images is within $\pm20^{\circ}$ from frontal position), background, age, gender, ethnicity, facial hair, hat, and glasses. All the images are manually labeled. GENKI-4K contains 2162 and 1838 smiling and non-smiling facial images, respectively. Sample images are shown in Fig.~\ref{Sample}.

\begin{table}[!t]
\caption{\label{tab1}GENKI-4K partition for cross-validation}
\centering
\begin{tabular}{|p{4cm}|l|l|l|l|}
\hline
Subset & 1 & 2 & 3 & 4 \\
\hline
Number of smile faces & 540 & 541 & 540 & 541 \\ 
\hline
Number of non-smile faces & 460 & 459 & 460 & 459 \\ 
\hline
\end{tabular}
\label{table1}
\end{table}

\begin{figure}
  \centering
    \includegraphics[width=\columnwidth]{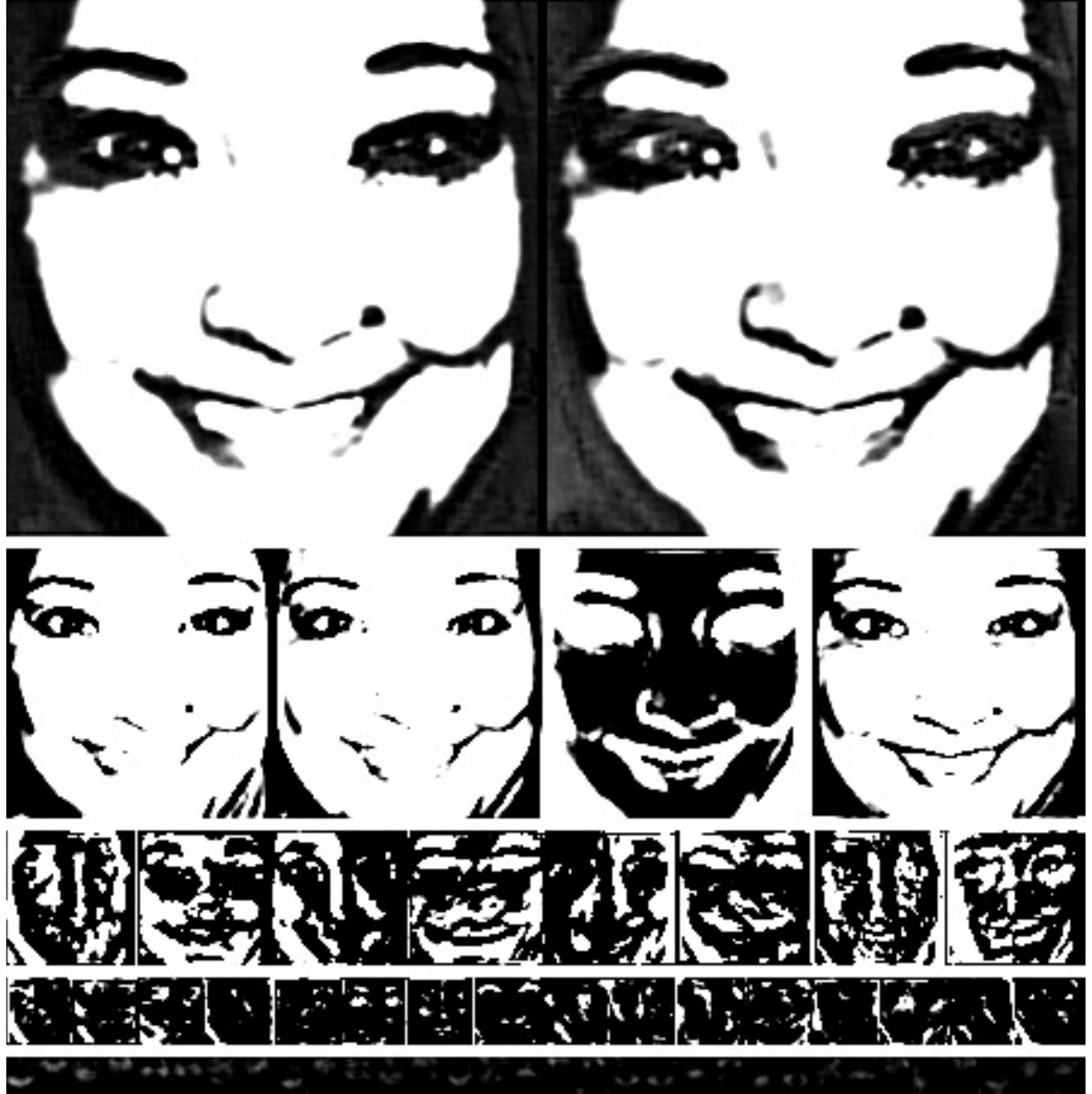}
  \caption{Random feature maps learned from CNN layers of a sample aligned face (refer to (\cite{Park15}) for a description of the VGG layers). First row: 2 feature maps learned from conv1\_2, each feature map of size $224\times224$. Second row: 4 feature maps learned from conv2\_2, each feature map of size $112\times112$. Third Row: 8 feature maps learned from conv3\_3, each feature map of size $56\times56$. Fourth row: 16 feature maps learned from conv4\_3, each feature map of size $28\times28$. Fifth row: 32 feature maps learned from conv5\_3, each feature map of size $14\times14$.}
  \label{conv_features}
\end{figure}

\subsection{Comparison with state-of-the-art methods}
To perform a fair comparison with state-of-the-art methods, the GENKI-4K dataset is first randomly divided into four subsets, having 1000 samples each, and then, those subsets are used for four-fold cross-validation. The number of smiling faces and non-smiling faces for each fold are shown in Table.~\ref{table1}. Each time, one subset is used for testing and the other three are used for training. The average detection rate and the standard deviation of the four-fold are reported as the final performance. All the images are preprocessed as described in Section \ref{preprocessing}.

\begin{figure}
  \centering
    \includegraphics[trim=1cm 0cm 0cm 0cm,clip=true,width=8.8cm]{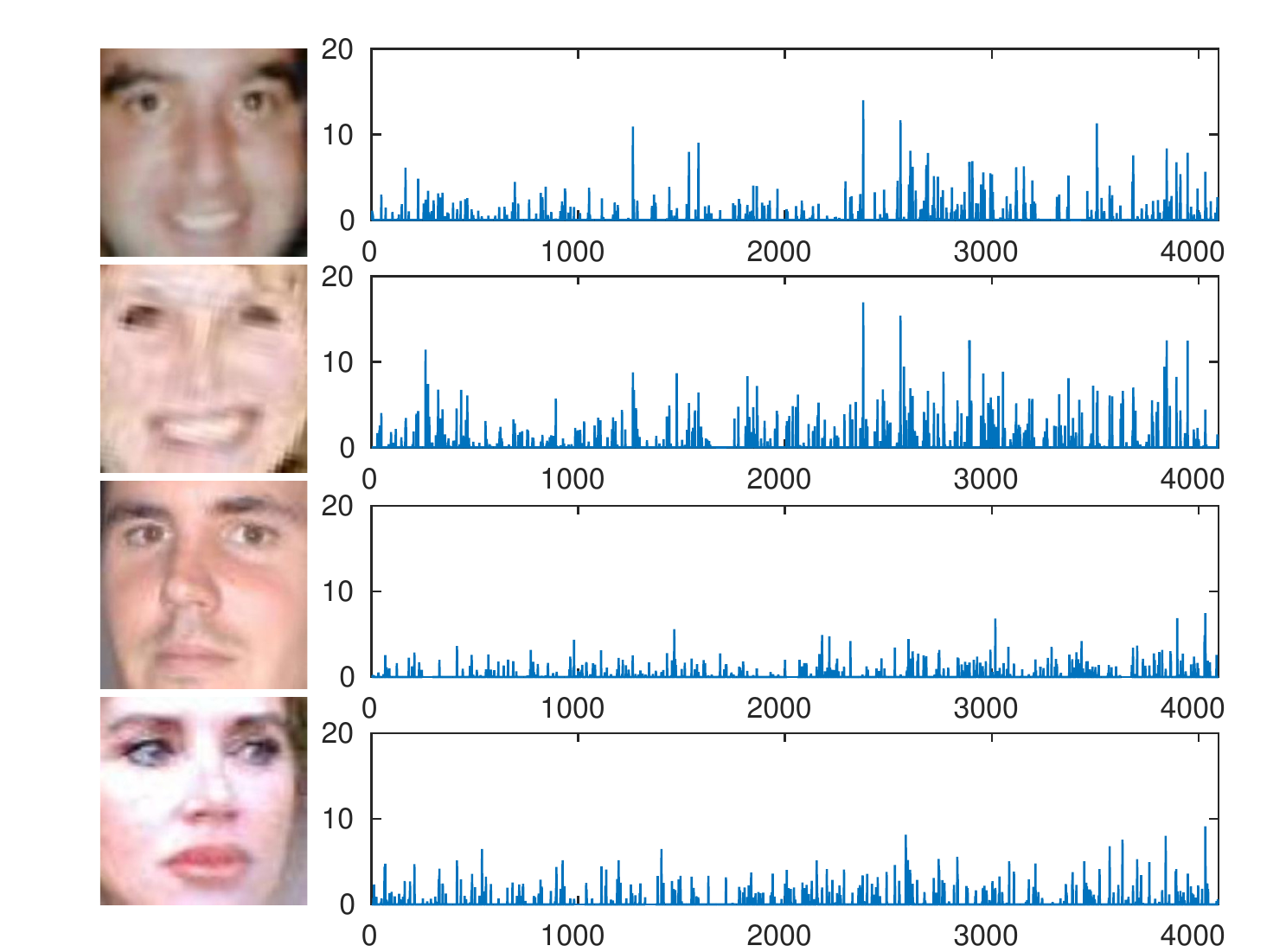}
  \caption{Features of fc7 layer for 2 smiling faces and 2 non-smiling faces. Left column: the aligned faces; right column: the values of features extracted after the fc7 layer, the horizontal and vertical axes correspond to the feature index and the feature values, respectively.}
  \label{fc7}
\end{figure}

As discussed in Section~\ref{fine-tuning}, the weights of all the convolutional layers are set the same as the VGG-face model weights because features extracted from CNN layers are generic features and because the VGG-face model was trained on 2.6M facial images. Sample feature maps from CNN layers of a smile image are shown in Fig.~\ref{conv_features}. It can be seen that low-level convolutional layers generally learn edges and outlines while the features become more abstract and sparse for upper layers. The first two and the last fully-connected layers are fine-tuned and trained from scratch on the GENKI-4K dataset, respectively. The features of the last fully-connected layer, fc7 (4096 features), are extracted and compared for 2 smiling and 2 non-smiling faces in Fig.~\ref{fc7}. As shown in the figure, features within the same class have similar feature values and trends, despite variations in gender, illumination and head pose. 

The number of weight and bias parameters of the original VGG-face model is around 138M. However, the number of trainable parameters of the fine-tuned VGG-face model is 120M since the parameters of all the convolutional layers are kept fixed during training. Despite having a reduction of only 13\% in the number of parameters, training takes only 30 minutes to converge at 1000 iterations. 

\begin{table*}[!t]
\caption{\label{tab2} Comparison with the state-of-the-art methods on the GENKI-4K database.}
\centering
\begin{tabular}{|l|l|l|l|}
\hline
Method & Features & Classifier & Accuracy(\%) \\
\hline
\multirow{2}{*}{~\cite{An15}} & LBP & ELM & 85.2 \\
                              & HOG & ELM & 88.2\\ 
\hline
\multirow{2}{*}{~\cite{Shan12}} & LBP & SVM & 87.1$\pm$0.76 \\
                              & Pixel comparison & AdaBoost & 89.7$\pm$0.45\\ 
\hline
\multirow{2}{*}{~\cite{Liu_Li_2012}} & HOG (labeled) & SVM & 91.8$\pm$0.97 \\
                              & HOG (labelled + unlabelled) & SVM & 92.3$\pm$0.81\\ 
\hline
\multirow{4}{*}{~\cite{Chen16}} & Raw pixels & SVM & 84.0$\pm$0.91 \\
                              & Raw pixels & AdaBoost & 80.0$\pm$0.76\\ 
                              & Learned features & SVM & 92.4$\pm$0.59\\ 
                              & Learned features & AdaBoost & 91.8$\pm$0.95\\ 
                              \hline
~\cite{V_Jain2014} & Guassian & SVM & 93.2$\pm$0.92 \\
\hline
\multirow{2}{*}{~\cite{Zhan15}} &CNN-Basic & Softmax & 93.6$\pm$0.47 \\
                              & CNN-2Loss & Softmax & 94.6$\pm$0.29\\
\hline
\multirow{5}{*}{~\cite{Gao:2016:NDG:2868729.2868956}} & HOG31 + GSS + Raw pixel & AdaBoost & 92.51$\pm$0.40 \\
                              & HOG31 + GSS + Raw pixel & Linear SVM & 94.28$\pm$0.60\\ 
                              & HOG31 + GSS + Raw pixel & Linear ELM & 94.21$\pm$0.35\\ 
                              & HOG31 + GSS + Raw pixel & Adaboost + Linear SVM & 94.56$\pm$0.62\\
                              & HOG31 + GSS + Raw pixel & Adaboost + Linear ELM & 94.61$\pm$0.53\\
\hline
The proposed method & Fine-tuned VGG-face model & Softmax & 95.38$\pm$0.52 \\
\hline
\end{tabular}
\label{table2}
\end{table*}

The performance of the fine-tuned VGG-face model is compared with state-of-the-art methods in Table~\ref{table2}. These methods are related to either the extraction of handcrafted features, such as histogram of oriented gradients (HOG) and local binary patterns (LBP), or to CNN models trained from scratch. The proposed fine-tuned VGG-face model outpeforms all the methods in Table~\ref{table2} in terms of classification accuracy and also exhibits a small variance. It is worth noting that most of the other methods use classifiers that are more sophisticated than the softmax classifier, such as SVM and extreme learning machine (ELM). However, the representational power of the features learned by CNN models reduces the need for using sophisticated classifiers.

\subsection{Impact of face alignment and color channels}
In a real-world scenario, facial landmarks may be hard to detect due to occlusion and illumination, resulting in  misaligned faces. Similarly, color images may not always be available, and we have to content ourselves with grayscale images. In this section, the impact of face alignment and color information in the performance of fine-tuned VGG-face models is evaluated.

Our approach to evaluate the impact of face alignment requires training and testing the model with the original unaligned GENKI-4K images using only face detection and cropping as preprocessing. On the other hand, to evaluate the impact of color information, the GENKI-4K images are first preprocessed as described in Section \ref{preprocessing}, then converted to grayscale and fed to the fine-tuned VGG model for training and testing. Note that since the input of the VGG-face model requires the three color channels, the same grayscale image is fed to the red, green, and blue channels to convert a single-channel image to a 3-channel one.

The same cross-validation partitioning described in Table~\ref{table1} is employed for this experiment and the results are compared with the classification accuracy attained using the original GENKI-4K images after cropping and face alignment. As expected, experimental results (Table~\ref{table3}) show that the models trained with the aligned RGB images perform better that the models trained with the unaligned and grayscale images. However, the decrease is small, less than 1\%, which means that the fine-tuned VGG-face model can achieve high classification accuracies even when the data exhibits high variability of head poses (within $\pm20^{\circ}$ from frontal position) and loss of color information.  

\begin{table}[!t]
\caption{\label{tab3} Classification accuracy (\%) on unaligned facial images and aligned grayscale facial images}
\centering
\begin{tabular}{|l|l|l|l|l|l|}
\hline
 & \small{Fold1} & \small{Fold2} & \small{Fold3} & \small{Fold4} & \small{Avg} \\
\hline
\small{RGB \& aligned} & 95.0 & 95.5 & 95.7 & 95.3 & 95.4\\ 
\hline
\small{RGB \& unaligned}& 94.8 & 95.2 & 95.1 & 94.1 & 94.8\\ 
\hline
\small{Grayscale \& aligned}& 94.4 & 95.1 & 94.9 & 95.0 & 94.9\\ 
\hline
\end{tabular}
\label{table3}
\end{table}

\subsection{Evaluation under image quality distortion}
Image distortions, such as image noise and blur, have demonstrated their power to fool a well-trained deep learning network (\cite{NIPS2014_5423},~\cite{Dodge2016}). In this section, the fine-tuned VGG-face model is evaluated on distorted images to evaluate its robustness.

Noise in the images may result from using low-quality camera senors in the real world. Noise is modeled as Gaussian noise added to each color channel of each pixel separately. The standard deviation of the noise is varied from 1 to 10 in steps of 1. 

\begin{figure*}
  \centering
    \includegraphics[width=\textwidth]{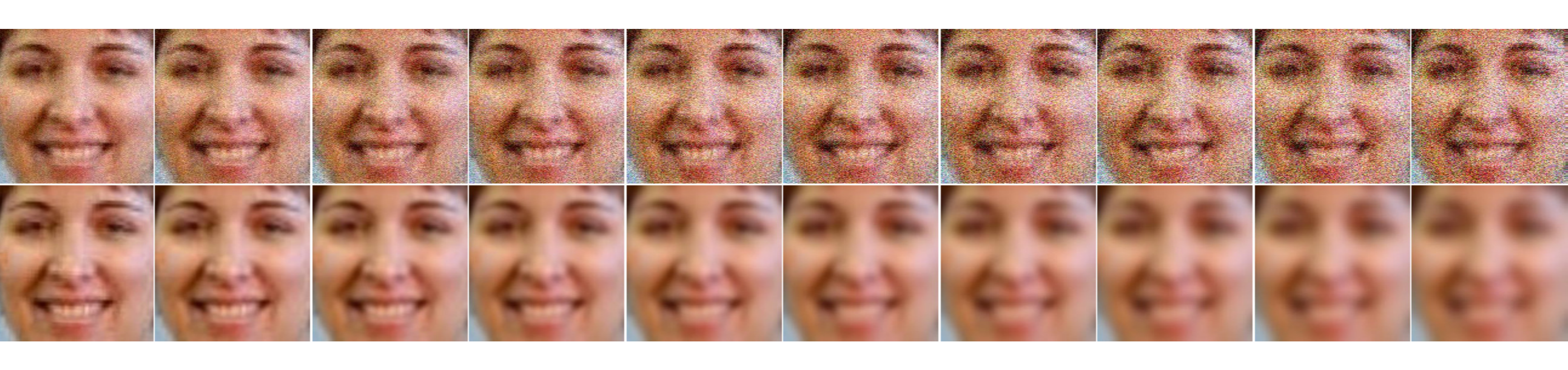}
  \caption{Sample images of noisy and blurred images. Top row: noisy images with the standard deviation of Gaussian noise varying from 1 to 10 in steps of 1. Bottom row: blurred images with the standard deviation of a Gaussian kernel varying from 1 to 10 in steps of 1.}
  \label{distort}
\end{figure*}

Blur may occur either because a camera is not focused properly on the target of interest or because the target is moving. A Gaussian kernel with varying standard deviation from 1 to 10 in steps of 1 is used to blur the images. The size of the filter window is set to 4 times the standard deviation. Sample images of noisy and blurred images under different Gaussian variations are shown in Fig.~\ref{distort}.

\begin{figure}
  \centering
    \includegraphics[width=\columnwidth]{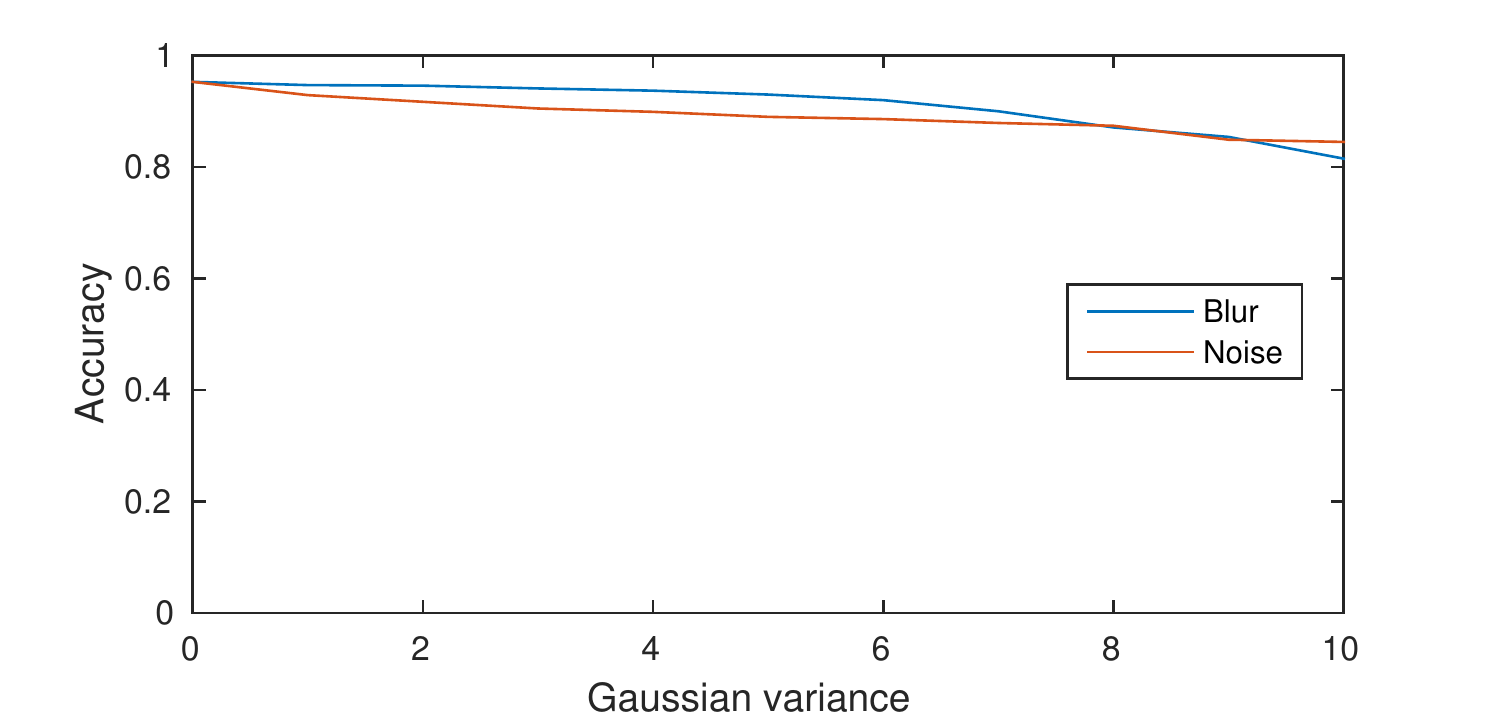}
  \caption{Performance evaluation of the fine-tuned VGG-face model in the presence of noise and blur artifacts.}
  \label{accuracy_noise_blur}
\end{figure}

The results in Fig.~\ref{accuracy_noise_blur} indicate that the fine-tuned VGG-face model is robust to noise and blur to some degree, given that the classification accuracy remains high (above 80\%) regardless of the image artifacts. It is worth noting that the accuracy of blurred images almost stays the same before the standard deviation of the Gaussian kernel reaches 6.

\section{Conclusions and discussion}
A smile detection method based on transfer learning was presented in this paper. Unlike previous research works which either perform feature extraction and classification separately or train a CNN from scratch, we leveraged the large labeled datasets and well-trained deep learning models in the face recognition field to generate a CNN model that achieves improved state-of-the-art on smile detection. The motivation behind this approach is that the labeled data on real-world smile detection datasets is scarce compared to the labeled data on real-world  face recognition datasets. For example, the GENKI-4K dataset has 4K images while the VGG-face recognition database has 2M images.

It was shown via experiments that the proposed method outperforms state-of-the-art methods for smile detection in terms of classification accuracy. However, transfer learning is not only important for increasing the classification accuracy, but it also reduces training time and leads to more robust systems because it exploits the high variability of large-scale datasets. For example, it was also shown that models also achieve high classification accuracy for smile recognition when fine-tuned and tested on artifact-corrupted images.

\bibliographystyle{IEEEbib}
\bibliography{RGCD2}

\end{document}